\documentclass[11pt]{article}

\usepackage[preprint]{acl}

\usepackage{times}
\usepackage{latexsym}

\usepackage[T1]{fontenc}

\usepackage[utf8]{inputenc}

\usepackage{microtype}

\usepackage{inconsolata}

\usepackage{amsmath,amssymb}
\usepackage{booktabs}
\usepackage{longtable}
\usepackage{array}
\usepackage{graphicx}
\usepackage{url}
\usepackage{placeins}

\title{Pair2Score: Pairwise-to-Absolute Transfer for LLM-Based Essay Scoring}
\author{
  \.{I}brahim R{\i}za Halla\c{c} \and Hasan O\u{g}ul \\
  Department of Computer Science and Communication \\
  Østfold University College, Halden 1757, Norway \\
  \texttt{\{ibrahim.r.hallac, hasan.ogul\}@hiof.no}
}

\begin{document}
\maketitle
\begin{abstract}
Many scoring applications require absolute predictions, while pairwise comparisons can provide a simpler learning objective. We present \textbf{Pair2Score}, a two-stage learning framework that transfers pairwise comparisons into absolute scoring with parameter-efficient LLaMA adaptation. Stage~1 trains a directional Siamese ranker on pairwise comparisons derived from absolute trait labels; Stage~2 trains an absolute predictor using configurable transfer strategies (warm-start and embedding-fusion variants). We evaluate on rubric-aligned Automated Essay Scoring (AES) traits (grammar, vocabulary, syntax) under a five-fold protocol that co-rotates held-out fold and random seed. At the trait level, the best-performing transfer variant improves quadratic weighted kappa (QWK) over an absolute-only baseline for all three traits. However, not all transfer configurations help: a one-epoch pairwise stage transfers more reliably than extended pairwise training, and transfer configuration---not just the inclusion of a pairwise stage---determines whether downstream scoring benefits.
\end{abstract}

\section{Introduction}

Absolute scoring is required in many evaluation settings, yet comparative judgments are often easier to collect and can carry stronger local information. This gap motivates a learning question: how to transfer pairwise evidence into reliable absolute scores.

We address this question with \textbf{Pair2Score}, a two-stage pipeline:
\begin{itemize}
\item \textbf{Stage 1 (relative):} learn directional pairwise ranking from document pairs.
\item \textbf{Stage 2 (absolute):} train an absolute scorer while reusing Stage~1 knowledge.
\end{itemize}

We use \textit{document} as a generic term; in our AES evaluation setting, each document is an essay. We evaluate on AES as an initial setting, but the formulation may apply to rubric-aligned scoring settings where comparative supervision can be derived from absolute labels. In our experiments, pairwise comparisons are derived from the same absolute trait labels used for Stage~2 evaluation, providing a local training objective for Stage~1.

\begin{figure*}[t]
\centering
\includegraphics[width=\textwidth,trim=16 14 16 12,clip]{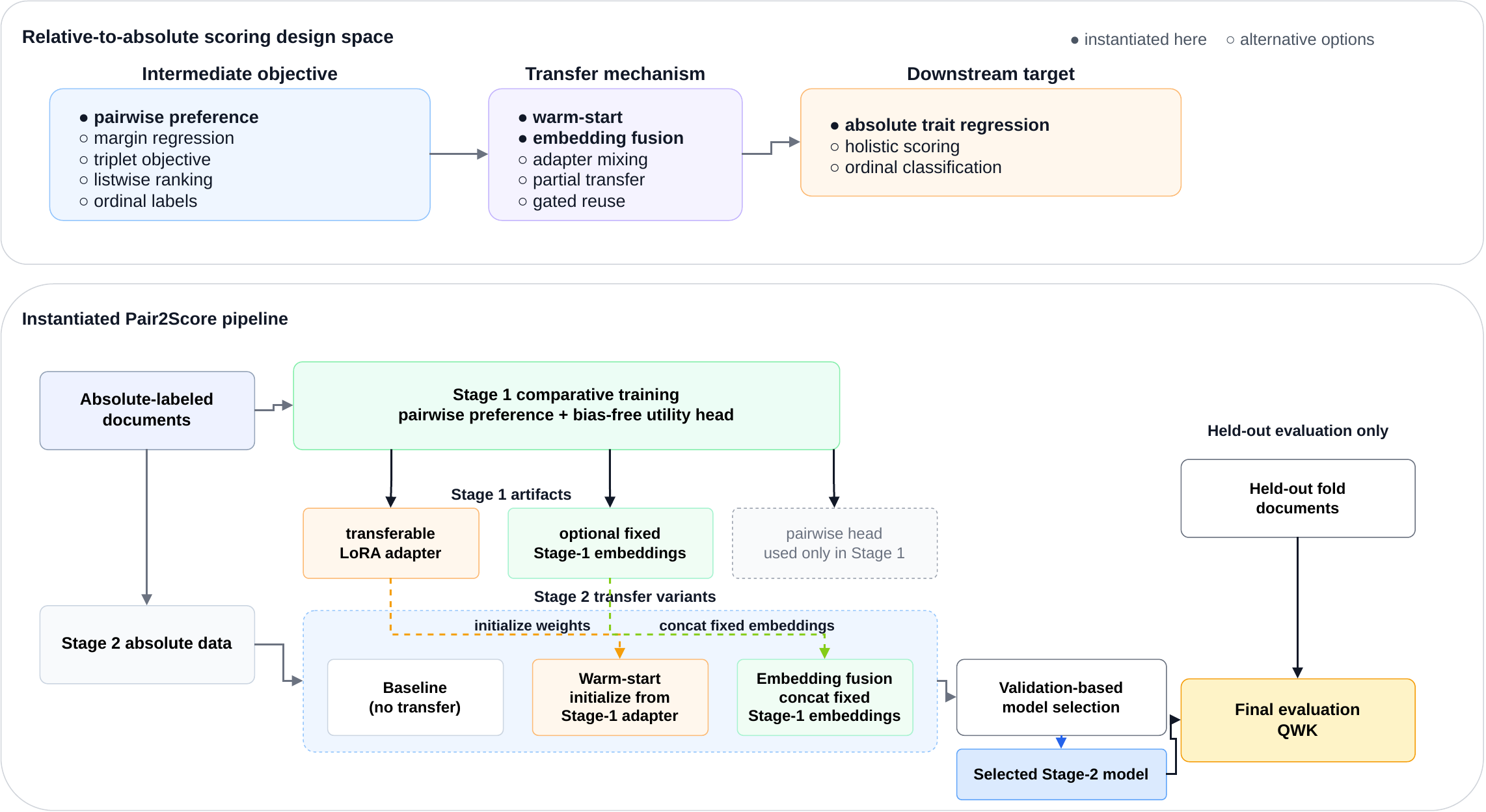}
\caption{Combined conceptual-and-protocol overview of Pair2Score. The top panel situates Pair2Score in the broader design space for relative-to-absolute scoring; filled markers denote the choices instantiated in this paper, and open markers denote alternative options. The bottom panel shows the instantiated two-stage protocol with Stage~1 artifacts, Stage~2 transfer variants, validation-based model selection, and held-out evaluation.}
\label{fig:overview}
\end{figure*}

In the LLM setting, a staged design is particularly natural: the same pretrained decoder-only backbone can be adapted to comparative and absolute objectives with lightweight task-specific heads and parameter-efficient updates. Pair2Score instantiates this with a LLaMA backbone, LoRA adapters, and small prediction heads for both stages (a bias-free linear utility head for Stage~1 and a regression head for Stage~2). We do not use prompting; scores are predicted directly from task-specific heads. Figure~\ref{fig:overview} situates Pair2Score in a broader relative-to-absolute scoring design space: our experiments instantiate one path through this space, using pairwise preference as the intermediate objective, warm-start and embedding fusion as transfer mechanisms, and absolute score regression as the downstream target.

We ask a single guiding question: \textbf{Can pairwise learning serve as an effective intermediate objective for rubric-aligned absolute essay scoring?} We study three design axes: whether to include pairwise training, how to transfer pairwise-trained information into the absolute scorer (warm-start vs embedding fusion), and how consistently effects hold across traits and cross-validation splits. One might expect that once Stage~1 reaches strong pairwise accuracy, Stage~2 would improve reliably. We evaluate this expectation directly in our experiments.

\subsection{Contributions}

\begin{enumerate}
\item We frame relative-to-absolute scoring transfer as a design space with three axes---intermediate objective, transfer mechanism, and downstream target (Figure~\ref{fig:overview})---and instantiate it as a two-stage pipeline (Pair2Score) with controlled transfer from a pairwise ranker to an absolute scorer. A co-rotated fold/seed evaluation protocol makes training stochasticity part of the measured evidence rather than averaging it away.
\item Across three essay-scoring traits and 15 fold-level evaluations, we find that pairwise ranking quality alone does not predict whether transfer improves absolute scoring; the choice of transfer mechanism and pairwise-stage budget matters at least as much.
\item The clearest axis is pairwise-stage duration: a single epoch of pairwise training transfers more reliably than extended training, suggesting that the pairwise stage contributes through early initialization rather than prolonged ranking optimization.
\end{enumerate}

\section{Related Work}

\subsection{Rubric-aligned scoring of open-ended responses (AES setting)}

Rubric-aligned scoring arises in many assessment workflows where open-ended written responses must be mapped to calibrated trait scores. Automated Essay Scoring (AES) is a standard evaluation setting for this problem, including trait-level prediction where separate dimensions (e.g., grammar, vocabulary, syntax) are modeled and evaluated \citep{ramesh_automated_2022}. In educational contexts, rubrics are widely used to structure such judgments, but reliability depends on rater training and rubric design \citep{jonsson_use_2007}. Some AES systems explicitly target agreement with human ratings during model training and selection \citep{chen_automated_2013,wang_automatic_2018}. Recent work continues to study AES with neural models and large language models (LLMs), including prompting-based scoring and feedback generation \citep{stahl_exploring_2024} and approaches that improve cross-prompt generalization \citep{bexte_increasing_2025}. Most systems optimize the absolute scoring objective directly; Pair2Score instead introduces a comparative intermediate objective and studies how that signal transfers into absolute trait scoring under a fixed protocol.

Comparative Judgment (CJ) is a complementary assessment paradigm where multiple assessors repeatedly compare essay pairs holistically, and latent quality scores are estimated from the resulting preferences, often with classical paired-comparison models such as Bradley--Terry \citeyearpar{bradley_rank_1952}. Beyond holistic ranking, comparative judgments also support pedagogical goals such as tracking whether a student's grammar or vocabulary improved between successive drafts. Recent work connects CJ and rubric-aligned scoring by (i)~predicting initial quality scores to warm-start CJ and reduce the number of required judgments \citep{de_vrindt_predicting_2024} and (ii)~predicting rubric scores to explain CJ-derived holistic scores \citep{de_vrindt_explaining_2025}. Pair2Score studies a related connection in a two-stage modeling setup: a comparative model is trained first, then transferred into an absolute scorer.

AES evaluation commonly reports quadratic weighted kappa (QWK), an agreement-style metric that accounts for the ordinal rubric scale \citep{ramesh_automated_2022}. Because QWK is not differentiable, many systems train with regression-style objectives and report QWK at evaluation time, while others incorporate agreement more directly into learning (e.g., by optimizing agreement-aware losses \citep{chen_automated_2013} or using reinforcement learning to optimize QWK \citep{wang_automatic_2018}). Since labels are provided by human raters, reported agreement also depends on label quality; recent work proposes metrics and protocols (e.g., PRMSE) to make this dependency explicit when comparing systems \citep{loukina_using_2020}.

\subsection{Pairwise ranking and comparative supervision}

Pairwise supervision is widely used in ranking and preference learning, including modern learning-to-rank objectives that operate on score differences \citep{chen_ranking_2009}. Related work also studies learning from discrete ordered labels and preference levels in ordinal prediction settings \citep{rennie_loss_2005}. In educational assessment, earlier and recent work revisit automated essay assessment from a relative perspective, including ranking-based AES approaches that model differences in writing quality between essays \citep{chen_automated_2014,leseanu_revisiting_2026}. Other AES work also combines pairwise and absolute supervision within a single-stage formulation, for example by jointly optimizing ranking and relative-score regression against reference essays \citep{xie_automated_2022}. Pair2Score studies a different point in this space: pairwise learning is used as a separate first stage whose learned representation is transferred into a downstream absolute scorer. In educational assessment, CJ is a domain-specific instance of this paradigm where pairwise comparisons are aggregated into latent quality scores \citep{de_vrindt_predicting_2024,de_vrindt_explaining_2025}.

\subsection{Intermediate objectives for absolute scoring}

Pairwise objectives are effective for learning relative orderings but do not directly yield calibrated absolute scores on discrete rubrics. Recent NLP work also shows that preference-based supervision can improve downstream absolute prediction without making ranking the end task: for example, auxiliary preference learning has been used to improve text classifiers while the deployed task remains ordinary classification \citep{kim_prefer_2023}. Pair2Score brings this perspective into rubric-aligned scoring by treating comparative supervision as a separate first stage whose learned representation is transferred into later absolute scoring.

\subsection{Parameter-efficient LLM adaptation and adapter transfer}

Parameter-efficient fine-tuning methods such as LoRA enable adaptation of large decoder-only LLM backbones with a small number of trainable parameters. Beyond efficiency, adapters provide a convenient carrier for staged transfer: a first-stage adaptation can be reused as initialization (warm-start) or as a fixed representation source (fusion-style transfer) in a second stage. Pair2Score uses LoRA as the transfer medium between the pairwise and absolute stages, comparing warm-start reuse versus embedding-fusion variants, while keeping the backbone and optimization protocol fixed to isolate transfer effects.
This design leverages parameter-efficient adaptation \citep{hu_lora_2021} and connects to work on adapter-based transfer and composition \citep{han_robust_2021,pfeiffer_adapterfusion_2021}.

\section{Method}

\subsection{Setup}

Given absolute labels $y$ on individual documents $x$, we construct ordered comparisons $(x_a, x_b)$ such that $y_a \ge y_b$. Our goal is to improve absolute prediction $f(x)$ by leveraging this comparative information.

Pair2Score has two stages:
\begin{enumerate}
\item Stage~1 trains a relative model on document pairs.
\item Stage~2 trains an absolute scorer on fold-based train/validation/test splits.
\end{enumerate}

We use Meta LLaMA-3.2-1B \citep{grattafiori_llama_2024} as the backbone in all experiments, with attention-mask mean pooling over final-layer token states (max length 512). We do not perform autoregressive generation; instead, we predict scores from pooled representations using lightweight heads (Stage~1: bias-free linear utility; Stage~2: 2-layer ReLU MLP with hidden size $d/2$). For parameter-efficient fine-tuning, we apply LoRA updates (rank 16, alpha 32, dropout 0.05) only to attention projections (q, k, v, o), keeping trainable parameters well under 1\% of the backbone. Only LoRA adapters and head parameters are updated during training.

\subsection{Stage 1: directional Siamese ranking}

\begin{figure*}[t]
\centering
\includegraphics[width=\textwidth]{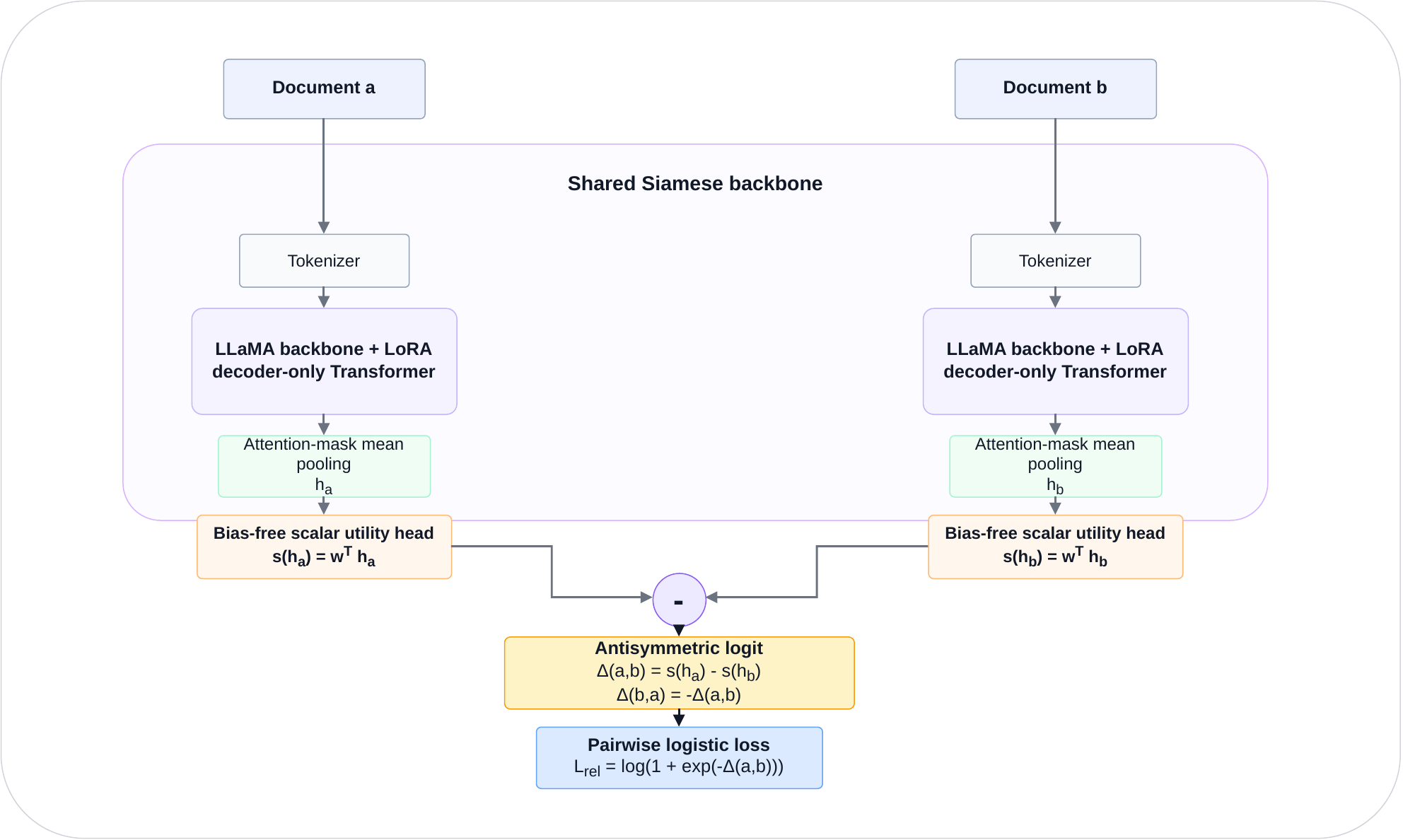}
\caption{Stage~1 directional Siamese objective. A shared LLaMA+LoRA backbone processes both documents; pooled representations $h_a, h_b$ are mapped to scalar utilities whose difference $\Delta(a,b)$ serves as the comparison logit (Section~3.2).}
\label{fig:siamese}
\end{figure*}

As illustrated in Figure~\ref{fig:siamese}, Stage~1 processes both documents in a Siamese setup with a shared LLaMA+LoRA backbone (and shared tokenizer), and compares them only through an antisymmetric utility difference.

Let $h_a, h_b$ be pooled hidden representations from a shared backbone. A bias-free linear utility head $s(\cdot)$ defines:
\[
\Delta(a,b)=s(h_a)-s(h_b).
\]
Thus,
\[
\Delta(b,a)=-\Delta(a,b),
\]
which enforces directional antisymmetry: given any two documents, if the model scores $a$ above $b$, reversing the input order automatically reverses the preference by exactly the same margin. This ensures that the comparative signal is consistent regardless of how the pair is presented.

The model is trained to produce $\Delta(a,b)>0$ for each training pair via pairwise logistic loss:
\[
\mathcal{L}_{\text{rel}}=\log\bigl(1+\exp(-\Delta(a,b))\bigr).
\]

We report Stage~1 pairwise accuracy as the fraction of comparisons where $\Delta(a,b) > 0$.

After Stage~1 training, we save the adapter weights and relative head parameters (and optionally pooled embeddings as diagnostics).

\subsection{Stage 2: absolute scoring with transfer}

Stage~2 trains an absolute scorer that predicts trait scores directly, optimizing L1 loss and reporting QWK. As shown in the bottom panel of Figure~\ref{fig:overview}, Stage~1 produces artifacts---adapter weights and optionally frozen embeddings---that Stage~2 can reuse. We compare two transfer mechanisms against an absolute-only baseline, and vary how much pairwise training each transfer setting receives:

\textbf{Baseline (absolute-only).} We attach a fresh LoRA adapter to the LLaMA backbone and train an absolute regression head directly on the absolute labels.

\textbf{Warm-start-only.} We first train a LoRA adapter using the pairwise objective. We then discard the pairwise head, initialize the absolute scorer from the resulting adapter weights, and fine-tune under the absolute objective. This tests whether pairwise supervision provides a useful initialization for absolute scoring.

\textbf{Fusion (embedding fusion).} In addition to warm-start initialization, we compute a Stage~1 embedding for each document once at the start of the absolute stage using the pairwise-trained backbone. During training, we concatenate this precomputed embedding with the current pooled representation before the regression head. This evaluates whether the pairwise stage contributes complementary features beyond initialization.

\textbf{Pairwise-stage settings (pair-set size and duration).} Smaller vs.\ larger refers only to the number of Stage~1 pairs: the smaller setting randomly subsamples 50\% of the available pool before pairing, while the larger setting uses the full pool. We also vary pairwise-stage duration by comparing the standard setting with a one-epoch pairwise stage; the absolute stage is unchanged.

\subsection{Optimization and monitoring}

We train both stages with early stopping and select model states using held-out validation data only. Stage~1 uses AdamW (batch 4, lr=2e-4, weight decay 0.01) for up to 10 epochs (1 epoch in the one-epoch Stage~1 setting), monitoring validation accuracy (patience 3, min\_delta 5e-4). Stage~2 uses AdamW (batch 4, lr=1e-4, weight decay 0.01) for up to 15 epochs, monitoring validation QWK (patience 4, min\_delta 5e-4). We do not use learning-rate schedulers. Test metrics in Section~5 come from the validation-selected model state.

To make stochasticity explicit (and to avoid relying on a single lucky seed), we fix a run-level random seed for each fold and keep it consistent across Stage~1 and Stage~2 within that run. We also keep training and numerical settings fixed across variants, so differences are attributable to the form of supervision and the transfer mechanism rather than changing training conditions.

\section{Data and Protocol}

\subsection{Dataset and targets}

We evaluate on the public Feedback Prize English Language Learning dataset \citep{kaggle_feedback_2021}, derived from the ELLIPSE corpus \citep{crossley_english_2024}, and focus on three rubric-aligned trait scores (grammar, vocabulary, syntax), selected a priori to keep the experimental comparison focused across distinct rubric dimensions. Each trait is treated independently: Stage~1 comparisons for a trait are derived from that trait's labels, and Stage~2 learns an absolute regressor for that trait. Trait labels lie in the range [1.0, 5.0] with 0.5 increments.

We perform five-fold evaluation using a fixed document-to-fold assignment (folds A--E) that is kept constant across all experiments. The fold map covers 3,911 documents (A=789, B=778, C=785, D=803, E=756).

Prompt identifiers are not available in the released version of the dataset; we generate pairs without prompt stratification.

\subsection{Stage 1 pair generation}

Stage~1 learns from document pairs created from the four training folds of each run. The held-out fold is never used for training or model selection in either stage; hyperparameter selection uses only validation data from the training folds.

For each trait and run, we split the available documents (from the training folds only) into Stage~1 train/validation/pair-test partitions at the document level (80/10/10). We then generate pairs independently within each partition, so no document (and no pair) can appear in more than one Stage~1 split. The Stage~1 pair-test split is used only for reporting Stage~1 accuracy/loss; Stage~2 evaluation uses the held-out fold.

\textbf{Pairing policy.} We construct pairs to balance coverage and informative comparisons. Each document participates in roughly five comparisons on average, and we target a minimum absolute label gap of 1.0 between paired documents. Pairs are generated in two phases:
\begin{enumerate}
\item \textbf{Coverage:} greedily create pairs so each document participates at least once.
\item \textbf{Fill:} add additional pairs by sampling within score-gap buckets ($\ge 3$, 2--3, 1--2) until the usage target is reached; if a split cannot satisfy the target under the constraints, we relax the usage cap to complete the pair set.
\end{enumerate}

We store each pair once in a canonical order based on the trait label (a: higher, b: lower), to avoid duplicating both (a,b) and (b,a) in the pair set.

In practice, the coverage phase prioritizes the score-gap constraint, but may include fallback pairs with smaller gaps to avoid dropping documents that cannot be paired under the constraint. The realized gap $d=\lvert y_a-y_b\rvert$ is stored for each pair.

\textbf{Pair-set size.} We consider two pair-set sizes with identical generation logic. The larger setting uses the full available pool within each Stage~1 split ($\approx$6k train pairs per trait/run); the smaller setting randomly subsamples 50\% of the pool before pairing ($\approx$3k train pairs), reducing Stage~1 cost.

\subsection{Fold+seed co-rotation protocol}

We run a five-fold evaluation. In each run, one fold is held out for testing in Stage~2, and the remaining four folds are used for Stage~1 and Stage~2 training (with an internal validation split for model selection). To capture nondeterminism from LLaMA fine-tuning, we co-rotate the training random seed with the held-out fold (A$\to$42, B$\to$48, C$\to$54, D$\to$60, E$\to$36). Stage~1 pair construction is deterministic and kept fixed per run/trait, so observed differences reflect learning and transfer mechanisms rather than resampled pairs.

\subsection{Evaluated variants and metrics}

We evaluate a baseline and eight transfer variants. The baseline trains the absolute scorer directly, with no pairwise stage. The eight transfer variants form a $2 \times 2 \times 2$ factorial crossing three design factors:
\begin{itemize}
\item \textbf{Transfer family}: warm-start (initialize the Stage~2 adapter from Stage~1) vs.\ fusion (warm-start + concatenate a precomputed Stage~1 embedding).
\item \textbf{Pair-set size}: small ($\approx$3k pairs) vs.\ large ($\approx$6k pairs) per trait and run (Section~4.2).
\item \textbf{Pairwise-stage duration}: standard (up to 10 epochs with early stopping) vs.\ one-epoch.
\end{itemize}

The primary evaluation metric is test QWK. Continuous predictions are clipped to the observed label range and discretized to the 0.5-step rubric grid before computing quadratic-weighted kappa.

\section{Results}

We report test QWK for the absolute scorer. For each trait and run, we compare the absolute-only baseline to the eight transfer settings defined in Section~4.4.

The sections below move from trait-level means to the full fold-level matrix and then to the contrast between best-case and average behavior across the evaluated settings.

\subsection{Trait-level means across folds}

\begin{table}[!ht]
\centering
\small
\begin{tabular}{@{}lccc@{}}
\toprule
Setting & Grammar & Vocab. & Syntax \\
\midrule
Baseline & 0.6789 & 0.6140 & 0.6474 \\
\midrule
\multicolumn{4}{@{}l}{\textit{Warm-start}} \\
\quad small, std & 0.6604 & 0.5959 & 0.6306 \\
\quad small, 1-ep & 0.6735 & 0.6152 & 0.6278 \\
\quad large, std & 0.6502 & 0.5970 & \textbf{0.6497} \\
\quad large, 1-ep & 0.6724 & 0.6000 & 0.6300 \\
\midrule
\multicolumn{4}{@{}l}{\textit{Fusion}} \\
\quad small, std & 0.6633 & 0.5914 & 0.6271 \\
\quad small, 1-ep & \textbf{0.6824} & \textbf{0.6197} & 0.6317 \\
\quad large, std & 0.6611 & 0.5920 & 0.6426 \\
\quad large, 1-ep & 0.6670 & 0.6019 & 0.6382 \\
\bottomrule
\end{tabular}
\caption{Trait-level mean QWK across five folds. Transfer variants are grouped by family; within each, rows vary pair-set size and pairwise-stage duration. Bold marks the best value per trait. Fold-level values are in Appendix Table~\ref{tab:fold_qwk}.}
\label{tab:trait_means}
\end{table}

Table~\ref{tab:trait_means} reports trait-level mean QWK across the five held-out folds under the co-rotated fold/seed protocol. Because these means aggregate over both fold differences and seed variation, they provide a compressed summary of a deliberately heterogeneous set of runs. Even under this aggregation, the best-performing transfer variant exceeds the absolute-only baseline for all three traits.

The fold-level results in Appendix Table~\ref{tab:fold_qwk} show a stronger pattern of improvement: underlined values mark transfer settings that exceed the baseline within a given trait-fold run. When the best transfer variant is selected for each trait-fold evaluation, pairwise transfer outperforms the baseline in \textbf{11 of 15} cases, with mean $\Delta$QWK = +0.012.

The winning variant differs by trait and fold, and no single setting dominates across runs. The next subsections examine which design choices---especially transfer mechanism and pairwise-stage duration---most reliably account for the observed improvements.

\FloatBarrier
\subsection{Transfer sensitivity: which configurations help?}

At the run level, pairwise transfer improves over the absolute-only baseline in 11 of 15 trait-fold evaluations, but these gains are distributed across multiple transfer settings rather than concentrated in a single dominant variant. Table~\ref{tab:best_variant_freq} summarizes how often each setting is the best performer: the most frequent winner is small fusion with a one-epoch pairwise stage (3 of 15 runs), while the remaining wins are spread across several other transfer configurations and the baseline.

\begin{table}[htbp]
\centering
\begin{tabular}{lr}
\toprule
Best variant & Count (of 15) \\
\midrule
Small, Fusion, 1-ep & 3 \\
Small, WS, 1-ep & 2 \\
Small, Fusion, std & 2 \\
Large, WS, std & 2 \\
Baseline & 4 \\
Other transfer & 2 \\
\bottomrule
\end{tabular}
\caption{Frequency of each variant being the best performer across the 15 trait-fold evaluations (Appendix Table~\ref{tab:fold_qwk}). WS = warm-start; 1-ep = one-epoch pairwise stage; std = standard duration. Rows with tied-best outcomes are assigned to one variant.}
\label{tab:best_variant_freq}
\end{table}

This dispersion is central to the interpretation of the results. Pairwise transfer can improve absolute scoring, but no single configuration consistently outperforms the others.

\FloatBarrier
\subsection{Does more pairwise training help?}

We isolate two design dimensions by holding all other variables fixed and comparing directly across 15 trait-fold runs. Table~\ref{tab:duration} reports the results: each row specifies which conditions are held fixed (pair-set size and transfer family in part~(a); transfer family and duration in part~(b)), while the remaining factor varies.

\begin{table}[t]
\centering
\small
{\setlength{\tabcolsep}{4pt}
\begin{tabular}{@{}lr*{3}{c}@{}}
\toprule
\multicolumn{5}{@{}l}{\textit{(a) One-epoch vs standard pairwise-stage duration}} \\
\midrule
Setting (held fixed) & $\Delta$QWK & \shortstack{1-ep\\wins} & \shortstack{Tie\\count} & \shortstack{Std\\wins} \\
\midrule
Small, Warm-start & +0.010 & 8 & 4 & 3 \\
Small, Fusion & +0.017 & 8 & 3 & 4 \\
Large, Warm-start & +0.002 & 6 & 4 & 5 \\
Large, Fusion & +0.004 & 5 & 7 & 3 \\
\bottomrule
\\[-2pt]
\toprule
\multicolumn{5}{@{}l}{\textit{(b) Larger vs smaller pair set}} \\
\midrule
Setting (held fixed) & $\Delta$QWK & \shortstack{Larger\\wins} & \shortstack{Tie\\count} & \shortstack{Smaller\\wins} \\
\midrule
Warm-start, standard & +0.003 & 7 & 3 & 5 \\
Warm-start, one-epoch & $-$0.005 & 7 & 2 & 6 \\
Fusion, standard & +0.005 & 9 & 1 & 5 \\
Fusion, one-epoch & $-$0.009 & 5 & 0 & 10 \\
\bottomrule
\end{tabular}
}
\caption{Direct comparisons across two design dimensions. Each row holds two dimensions fixed and varies a third. Part~(a): one-epoch vs standard duration; part~(b): larger vs smaller pair set. $\Delta$QWK is the mean difference across 15 runs (positive favors the first condition). Tie counts indicate the number of trait-fold runs in which the two compared conditions obtain the same test QWK.}
\label{tab:duration}
\end{table}

\paragraph{Duration (part a).} In the small-pair-set families, the one-epoch variant wins or ties in 11--12 of 15 runs for warm-start and fusion alike. The advantage narrows for the larger pair set but mean $\Delta$QWK remains non-negative in all four rows.

\paragraph{Pair-set size (part b).} Across the four comparisons, mean $\Delta$QWK values are small and mixed in sign, and win counts do not indicate a consistent advantage for either pair-set size.

Duration shows a clear pattern; pair-set size does not. This suggests that the pairwise stage's value lies in early initialization rather than in prolonged ranking optimization or larger training sets.

\subsection{Variation across folds and transfer settings}

Fold-level variability is expected under the co-rotated fold/seed protocol, and we report it as part of the evidence rather than filtering it out through single-split reporting.

In Table~\ref{tab:fold_qwk}, baseline test QWK across the 15 trait-fold runs ranges from 0.588 to 0.709 (std 0.033). When we instead select the best-performing transfer variant for each run, the resulting scores range from 0.623 to 0.707 (std 0.027). These ranges demonstrate that substantial fold and seed variation persists even under otherwise fixed training conditions.

\section{Analysis}

A natural question is whether a stronger Stage~1 ranker reliably translates into better absolute scoring in Stage~2. We test this by pairing each transfer-setting Stage~2 run with its corresponding Stage~1 run (same trait, held-out fold/seed, pair-set size, transfer family, and pairwise-stage budget) and then measuring association between Stage~1 ranking metrics and Stage~2 test QWK across all transfer-setting runs ($n=120$; 3 traits $\times$ 5 folds $\times$ 8 transfer settings).

Stage~1 metrics are computed on the Stage~1 pair-test split, while Stage~2 QWK is computed on the held-out fold.

We report Pearson correlation (linear association) and Spearman correlation (rank association). Values near 0 indicate weak association:
\begin{itemize}
\item Overall Stage~1 pair-test accuracy vs Stage~2 test QWK: \textbf{+0.05}
\item By trait (Pearson, Stage~1 pair-test accuracy $\rightarrow$ Stage~2 test QWK): grammar \textbf{$-$0.23}, vocabulary \textbf{$-$0.06}, syntax \textbf{$-$0.10}
\end{itemize}

To check whether other Stage~1 diagnostics are more predictive, we repeated the analysis for validation accuracy, validation loss, and selected best epoch. Under both Pearson and Spearman, all associations with Stage~2 QWK remain weak (Pearson range: $-$0.21 to +0.11; Spearman range: $-$0.17 to +0.10; $n = 120$). The directions are consistent with expectation---validation accuracy weakly positive, validation loss weakly negative, training duration weakly negative---but none is large enough to be informative.

These results indicate that, within the range of Stage~1 performance observed here, downstream performance depends more strongly on transfer configuration and held-out-fold differences than on marginal gains in Stage~1 ranking accuracy. In other words, higher standalone Stage~1 ranking accuracy does not reliably translate into better absolute scoring in Stage~2 under the transfer settings evaluated here.

We do not interpret this as evidence that pairwise ranking quality is inherently irrelevant to absolute scoring. A stronger comparative representation should, in principle, provide a richer basis for downstream prediction. The weak association we observe more likely reflects the limited expressiveness of our transfer pathways (LoRA warm-start and single-vector embedding fusion), which may not preserve the comparative information in a form that the absolute scorer can fully exploit. Designing transfer mechanisms that close this gap---for example through progressive distillation, richer intermediate representations, or jointly learned transfer layers---is an open direction that our results motivate.

\section{Conclusion}

Pair2Score treats comparative supervision as a separate intermediate stage for LLM-based absolute essay scoring and places this choice in a broader relative-to-absolute scoring design space. In our AES setting, a pairwise stage can improve downstream absolute scoring, but the effect depends on how comparative information is transferred into the scorer. Among the axes we study, pairwise-stage duration shows the clearest pattern: a one-epoch pairwise stage transfers more reliably than extended training, suggesting that the pairwise stage contributes primarily through early initialization rather than through prolonged ranking optimization. The central result is therefore not simply that pairwise training helps, but that intermediate objective, transfer pathway, and downstream target have to be considered together. This makes relative-to-absolute transfer a modeling question in its own right. Although we study scoring in this paper, the same design space may also matter for later assessment workflows, including feedback-oriented systems that use comparative signals before producing rubric-aligned outputs.

\section*{Acknowledgments}

This work was supported by the AI4AfL project, funded by the Research Council of Norway (grant number 326607).

\section*{Limitations}

Our evaluation instantiates only one path through the broader relative-to-absolute scoring design space in Figure~\ref{fig:overview}: pairwise preference as the intermediate objective, warm-start and embedding fusion as transfer mechanisms, and absolute score regression as the downstream target, all in a text-based AES setting with three rubric-aligned traits and a fixed pair-generation policy. The present study is therefore a focused test of one relative-to-absolute pathway rather than a full account of the design space.

For practical use, this leaves open how the same pattern would extend to other intermediate objectives, transfer pathways, prompts, and assessment workflows. An important next step is cross-dataset relative-to-absolute transfer, where Stage~1 is trained on one or more source datasets and Stage~2 is evaluated on a separate target dataset. Such settings would test whether the comparative signal learned in Stage~1 generalizes beyond a single dataset and rubric configuration, and may in turn motivate lighter transfer mechanisms tailored to this setting. Settings with independently collected comparative judgments, prompt-aware scoring, or feedback-oriented outputs would likewise require separate validation.

We evaluate only two transfer mechanisms (warm-start and embedding fusion) with a single backbone (LLaMA-3.2-1B). We do not compare against external AES baselines; all comparisons are internal to the Pair2Score framework. Results on additional backbones, datasets, and transfer mechanisms are needed before broader conclusions can be drawn.

\section*{Ethics Statement}

Our experiments use the publicly available Feedback Prize / ELLIPSE corpus \citep{kaggle_feedback_2021,crossley_english_2024} for research purposes only, and we do not attempt to re-identify individual writers. Although pairwise supervision may reduce some noise relative to direct score regression, it does not remove broader risks of unfair or misleading automated scoring; any practical use should therefore include human oversight and clear communication of model limitations.

\appendix

\section{Fold-level QWK matrix}
\label{app:fold_qwk}

Table~\ref{tab:fold_qwk} reports fold-level test QWK for each trait and held-out fold. Columns correspond to the baseline and the eight transfer settings defined in Section~4.4.

\begin{table*}[h]
\centering
\scriptsize
\resizebox{\linewidth}{!}{%
\begin{tabular}{lllllllllll}
\toprule
Trait & Fold & Baseline & SWS & SWS-r1 & SF & SF-r1 & LWS & LWS-r1 & LF & LF-r1 \\
\midrule
grammar & E & 0.6556 & 0.6116 & 0.6421 & 0.6073 & \underline{\textbf{0.6738}} & 0.6009 & \underline{0.6657} & 0.6311 & \underline{0.6597} \\
 & A & \textbf{0.7085} & 0.6816 & 0.6964 & 0.6896 & 0.7065 & 0.6941 & 0.7010 & 0.6910 & 0.6942 \\
 & B & 0.6761 & 0.6380 & \underline{\textbf{0.6812}} & 0.6645 & 0.6524 & 0.6506 & 0.6506 & 0.6672 & 0.6646 \\
 & C & 0.6563 & \underline{0.6960} & \underline{0.6960} & \underline{\textbf{0.6980}} & \underline{\textbf{0.6980}} & \underline{0.6576} & \underline{0.6576} & \underline{0.6772} & \underline{0.6772} \\
 & D & \textbf{0.6982} & 0.6749 & 0.6520 & 0.6573 & 0.6813 & 0.6478 & 0.6869 & 0.6391 & 0.6391 \\
\addlinespace
vocabulary & E & 0.6151 & 0.5985 & \underline{\textbf{0.6409}} & 0.5982 & 0.5909 & 0.5985 & 0.5706 & 0.5500 & 0.5795 \\
 & A & 0.6311 & 0.6157 & \underline{0.6338} & 0.5724 & \underline{\textbf{0.6349}} & 0.6151 & 0.6251 & 0.5724 & 0.6289 \\
 & B & 0.6226 & 0.5653 & 0.5653 & \underline{\textbf{0.6231}} & \underline{\textbf{0.6231}} & 0.5653 & 0.6000 & 0.5975 & 0.5975 \\
 & C & 0.6131 & 0.6116 & \underline{0.6158} & 0.5900 & \underline{0.6196} & \underline{0.6177} & \underline{0.6158} & \underline{\textbf{0.6397}} & \underline{\textbf{0.6397}} \\
 & D & 0.5882 & \underline{0.5886} & \underline{0.6201} & 0.5734 & \underline{\textbf{0.6299}} & \underline{0.5886} & \underline{0.5886} & \underline{0.6002} & 0.5638 \\
\addlinespace
syntax & E & \textbf{0.6418} & 0.6223 & 0.6223 & 0.6223 & 0.6223 & 0.6251 & 0.6387 & 0.6251 & 0.6251 \\
 & A & 0.6497 & \underline{\textbf{0.6881}} & \underline{0.6778} & \underline{\textbf{0.6881}} & \underline{0.6778} & \underline{0.6769} & 0.6443 & \underline{0.6769} & 0.6443 \\
 & B & 0.6217 & 0.6169 & 0.6066 & 0.6150 & 0.6066 & \underline{\textbf{0.6275}} & \underline{\textbf{0.6275}} & 0.6169 & \underline{\textbf{0.6275}} \\
 & C & \textbf{0.6895} & 0.6313 & 0.6378 & 0.6313 & 0.6698 & 0.6698 & 0.6452 & 0.6452 & 0.6452 \\
 & D & 0.6344 & 0.5943 & 0.5943 & 0.5788 & 0.5821 & \underline{\textbf{0.6491}} & 0.5943 & \underline{\textbf{0.6491}} & \underline{\textbf{0.6491}} \\
\bottomrule
\end{tabular}
}
\caption{Fold-level test QWK for each trait and held-out fold. Bold marks the highest value in each row; underlining marks transfer settings that exceed the baseline in the same row. Column codes: S/L = small/large pair set; WS/F = warm-start/fusion; -r1 = one-epoch pairwise stage.}
\label{tab:fold_qwk}
\end{table*}

\section{Reproducibility}

All training runs enforce full numerical determinism: global seeds are fixed for all random sources, GPU non-determinism is disabled, and separate generator seeds isolate each data-loader split so that shuffling in one partition does not affect another. The global seed is co-rotated with the held-out fold (Section~4.3). Code, experiment configurations, determinism utilities, and fold-level results are available at \url{https://github.com/irhallac/pair2score}.

\section{Implementation details}

Our implementation follows the paper structure: a Stage~1 relative training module, a Stage~2 absolute training module, and a pipeline wrapper that runs the two stages under a single configuration. Data preparation consists of (i)~creating a fixed fold map and (ii)~generating per-run pair caches restricted to the training folds, with metadata to trace how each cache was constructed.

\subsection{Computing information}

We run experiments in Python using PyTorch and the Hugging Face Transformers stack, training on a single NVIDIA GPU per run. We keep the software stack fixed: Ubuntu 20.04.6 LTS with Python 3.10.18 and PyTorch 2.4.0+cu121 (CUDA 12.1; cuDNN 9.1), using Transformers 4.57.1, PEFT 0.15.0, and Accelerate 0.33.0. Representative GPU types used in our runs include NVIDIA Quadro RTX 8000 (48GB) and NVIDIA A100 80GB PCIe.

\subsection{Hyperparameters (summary)}

We use AdamW and early stopping in both stages, selecting model states using validation data only (see \S3.4).

\begin{center}
\small
\begin{tabular}{>{\raggedright\arraybackslash}p{0.20\linewidth}>{\raggedright\arraybackslash}p{0.45\linewidth}>{\raggedright\arraybackslash}p{0.25\linewidth}}
\toprule
Stage & Training setup & Model selection \\
\midrule
Stage~1 (relative) & AdamW; batch 4; lr 2e-4; weight decay 0.01; max 10 epochs (1 for one-epoch variants) & validation accuracy (patience 3) \\
Stage~2 (absolute) & AdamW; batch 4; lr 1e-4; weight decay 0.01; max 15 epochs & validation QWK (patience 4) \\
\bottomrule
\end{tabular}
\end{center}

Other fixed settings:
\begin{itemize}
\item Max sequence length: 512; pooling: attention-mask mean pooling over final-layer token states.
\item LoRA: rank 16, alpha 32, dropout 0.05; target modules: attention projections (q, k, v, o).
\end{itemize}

\bibliography{references}
\end{document}